%% file: AnonymousSubmission2027.tex
\documentclass[letterpaper]{article} 
\usepackage{aaai2027}  
\usepackage[hyphens]{url}  
\usepackage{graphicx} 
\usepackage{natbib}  
\usepackage{caption} 
\usepackage{algorithm}
\usepackage{algorithmic}

\usepackage{newfloat}
\usepackage{listings}
\DeclareCaptionStyle{ruled}{labelfont=normalfont,labelsep=colon,strut=off} 
\floatstyle{ruled}
\newfloat{listing}{tb}{lst}{}
\floatname{listing}{Listing}

\usepackage{booktabs}
\usepackage{multirow}
\usepackage{bm}
\usepackage{amsmath}
\usepackage{amssymb}
\usepackage{tabularx}
\usepackage{array}
\usepackage{tcolorbox}

\newtheorem{definition}{Definition}

\newtheorem{lemma}{Lemma}
\newtheorem{theorem}{Theorem}

\definecolor{commentcolor}{RGB}{70,120,70}
\definecolor{chronoyellow}{HTML}{FFF8D8}
\definecolor{discussionback}{HTML}{EEF7EE}
\definecolor{discussionframe}{HTML}{333A33}

\newtcolorbox{chronobox}{
  colframe=black!65,
  colback=yellow!5,
  boxrule=0.9pt,
  arc=4mm,
  left=10pt,
  right=10pt,
  top=7pt,
  bottom=7pt,
  boxsep=0pt
}

\title{CertVLA: Certified Defense against Physical Visual Attacks for Vision-Language-Action Models}
\author{
    Hui Lu, Zhijie Peng, Yuqi Lin, Zaijia Yang, Jiaming He, Shuhan Ye, Yi Yu\corresponding, Hanwei Zhu, Bingquan Shen, Alex Kot, Xudong Jiang 
}
\affiliations{

}

\begin{document}

\maketitle

\begin{abstract}
Vision-Language-Action (VLA) policies are vulnerable to localized physical perturbations, yet existing certified patch defenses target discrete labels and cannot directly certify continuous, temporally correlated actions. We introduce CertVLA, a certified defense for closed-loop VLA control under bounded patch and texture attacks. CertVLA proposes a calibrated region of behaviorally consistent actions, while deterministic covering masks ensure that at least one checked prediction is attack-free. Specifically, CertVLA normalizes action disagreement by the benign variation of each mask pair and accepts a single-mask anchor only when it remains consistent under every second mask. It then calibrates the resulting max-min-max episode score to provide finite-sample clean coverage. Conjoining query-level decisions extends the action certificate to the complete closed-loop rollout. Furthermore, we prove that against any adaptive attacker satisfying the bounded-support threat model, every rollout certified by CertVLA executes only action chunks consistent with attack-erased clean predictions. Under dual-mask rollout correctness, this consistency certificate further guarantees task success. The certificate is independent of patch content, generation method, and physical transformation. Experiments in simulation and the real world demonstrate the empirical and certified effectiveness of CertVLA against patch attacks, with additional simulation validation on texture attacks.

\end{abstract}


\section{Introduction}

Vision-Language-Action (VLA) models connect visual perception and language
instructions directly to robot control, enabling general-purpose policies for
diverse manipulation tasks~\cite{zitkovich2023rt2,octo2024,black2024pi0,black2025pi05}.
This tight perception-action coupling also creates a physical attack surface:
a localized patch in the scene or an adversarial texture on an object can
persist across viewpoints and corrupt an entire closed-loop trajectory.
Recent attacks have demonstrated such failures across tasks, VLA
architectures, and sim-to-real settings~\cite{wang2025vlaattack,lu2026robots,chen2026tex3d}.
Existing empirical defenses improve VLA robustness through robust training,
decoupled robustness learning, safety-constrained optimization, and
noise-filtering modules
~\cite{guo2025robustvla,xie2026strongvla,fu2026stablevla}, while patch-specific defenses in vision typically localize, mask, or restore suspicious regions~\cite{tarchoun2023jedi,xu2023patchzero,jing2024pad,yankelev2026antistyler}.
For safety-critical deployment, however, empirical robustness alone is insufficient,
as resistance to tested attacks does not establish robustness to all attacks
within the threat model~\cite{cohen2019certified,levine2020robustness,xiang2022patchcleanser}.
A certified defense should additionally determine when the executed actions
are provably protected against every admissible attack.

Existing certified patch defenses provide a natural starting point for this
goal. Prior work has developed deterministic ablation, restricted receptive
fields, and masking-based approaches to certify discrete predictions against
bounded patch attacks
~\cite{levine2020derandomized,xiang2021patchguard}.
Extending this paradigm to VLAs, however, requires addressing fundamental
differences between discrete perception and embodied control. \textit{First}, VLAs typically
produce continuous action chunks, for which exact agreement among masked
outputs is generally too restrictive and a measurable notion of action
consistency is needed. \textit{Second}, although covering-based masking~\cite{lyu2026certmask,xiang2022patchcleanser} can handle an unknown patch location,
its discrete recovery criterion does not directly apply to continuous actions.
The defender must instead determine whether a recovered action remains within
a calibrated region of an attack-erased reference without knowing which mask
removes the attack. \textit{Third}, VLA actions
are executed in closed loop. A certificate at one policy query only constrains
the current action chunk, while its execution changes the state and observations
encountered at subsequent queries. Certification must therefore extend from
individual predictions to the sequence of decisions made along the actual
rollout. 
Hence, a
research question arises naturally:

\begin{chronobox}
\small
\textit{How can we certify continuous VLA actions against unknown bounded physical perturbations and extend the guarantee from individual policy queries to closed-loop task execution?} 
\end{chronobox}

In response, we introduce \emph{CertVLA}, a model-agnostic certified defense for continuous
VLA control under bounded patch and texture attacks. CertVLA uses an
$\mathcal R$-covering mask family~\cite{xiang2022patchcleanser} so that, for any admissible support,
at least one evaluated mask completely erases the corrupted region. Because
different mask positions induce different benign action changes, we normalize
each anchored dual-mask disagreement by a position-specific clean scale and
calibrate a shared tolerance with held-out clean episodes. The resulting
max-min-max score exactly represents the required
$\forall$-query--$\exists$-anchor--$\forall$-second-mask decision. At
deployment, an early-stopped recovery rule rejects a row as soon as it cannot
pass and returns a certified anchor only if all second-mask checks pass.

This construction yields a deliberately layered guarantee. Whenever CertVLA
certifies a query, the returned action chunk lies in a calibrated
clean-consistency region of a prediction for which the attack has been erased.
The statement holds for arbitrary patch content and adaptive attack generation, provided that the attack support can be fully enclosed by a fixed-size square region. Conjoining query-level flags extends
the result to the complete executed rollout, even when the admissible patch
location changes over time. Under an explicit dual-mask rollout-correctness
condition, the same certificate further implies terminal task success. 

The contributions of our paper are summarized as follows:
\begin{itemize}
    \item We formulate certified defense for continuous closed-loop VLA control
and introduce a calibrated action-consistency region beyond exact label
agreement.
    \item We develop an anchored dual-mask defense that couples position-aware
normalization with a covering-based recovery rule for certifiable action
selection.
    \item We prove query- and episode-level consistency guarantees against any
    bounded-support adaptive attacker, and isolate the precise additional
    condition under which this observable certificate guarantees task
    completion.
    \item Extensive experiments across multiple VLAs, attack types, and deployment settings
demonstrate the broad applicability of CertVLA.
\end{itemize}

\input{sections/certvla_core}

\section{Experiments}

\paragraph{Dataset \& threat models.}
We use LIBERO~\cite{liu2023libero}, a simulated manipulation benchmark, and
evaluate its Spatial, Object, Goal, and Long suites with 10 tasks for each, including spatial relations, object interaction, goal-conditioned control,
and multi-stage execution. We report results for four VLA policies: OpenVLA~\cite{kim2025openvla}, OpenVLA-OFT~\cite{kim2025fine}, $\pi_0$~\cite{black2024pi0}, and $\pi_{0.5}$~\cite{black2025pi05}, across all suites. Attacks are physical patches~\cite{wang2025vlaattack, lu2026robots}
and adversarial object textures~\cite{chen2026tex3d}. In both cases, the visible image-plane corruption must satisfy the bounded-support threat model.

\paragraph{Evaluation metrics \& details.}
For each model, task suite, and attack setting, we evaluate $N$ closed-loop episodes per suite.
\emph{Defense} reports the defended task success rate,
$N^{-1}\sum_{e=1}^{N}\mathbb{I}[S_e^{\rm def}=1]$, where $S_e^{\rm def}$
denotes task success under defense. \emph{Certified} reports
$N^{-1}\sum_{e=1}^{N}\mathbb{I}[S_e^{\rm def}=1\land C_e=\mathrm{true}]$,
where $C_e$ means all policy queries in episode $e$ pass the consistency test.
Thus, Certified counts only defended successes with an end-to-end certificate.
We set $\beta=0.95$, $\alpha=0.5$, and $\epsilon=10^{-8}$ for all main
results.
See \textit{Supplementary} for more experimental details.

\paragraph{Physical Experiment Setting.}
For real-world experiments, we employ a dual-arm Piper robotic platform. Each arm is controlled using a 7-dimensional end-effector action. We use one arm to implement the pick-and-place task. Visual observations are provided by two RGB cameras: one Intel RealSense D435 camera offering a third-person view and one Intel RealSense D405 camera mounted on the wrist. Our policy uses these synchronized RGB observations together with the arm proprioceptive state. We use standardized objects and workspace layouts for systematic quantitative evaluation, with the task repeated for 10 independent trials. 
\begin{figure}[t]
    \centering
    \small
        \begin{tabularx}{0.9\linewidth}{
            @{}l*{4}{>{\centering\arraybackslash}X}@{}
        }
            \toprule
            Model & Clean & Attack & Defense & Certified \\
            \midrule
            $\pi_{0.5}$ & 90 & 40 & 60 & 30 \\
            \bottomrule
        \end{tabularx}
        \captionof{table}{Results for
        physical patch attacks on the real robot.}
        \label{tab:real-patch-results}
\end{figure}
\begin{figure*}[t]
\centering
    \includegraphics[width=0.87\linewidth]{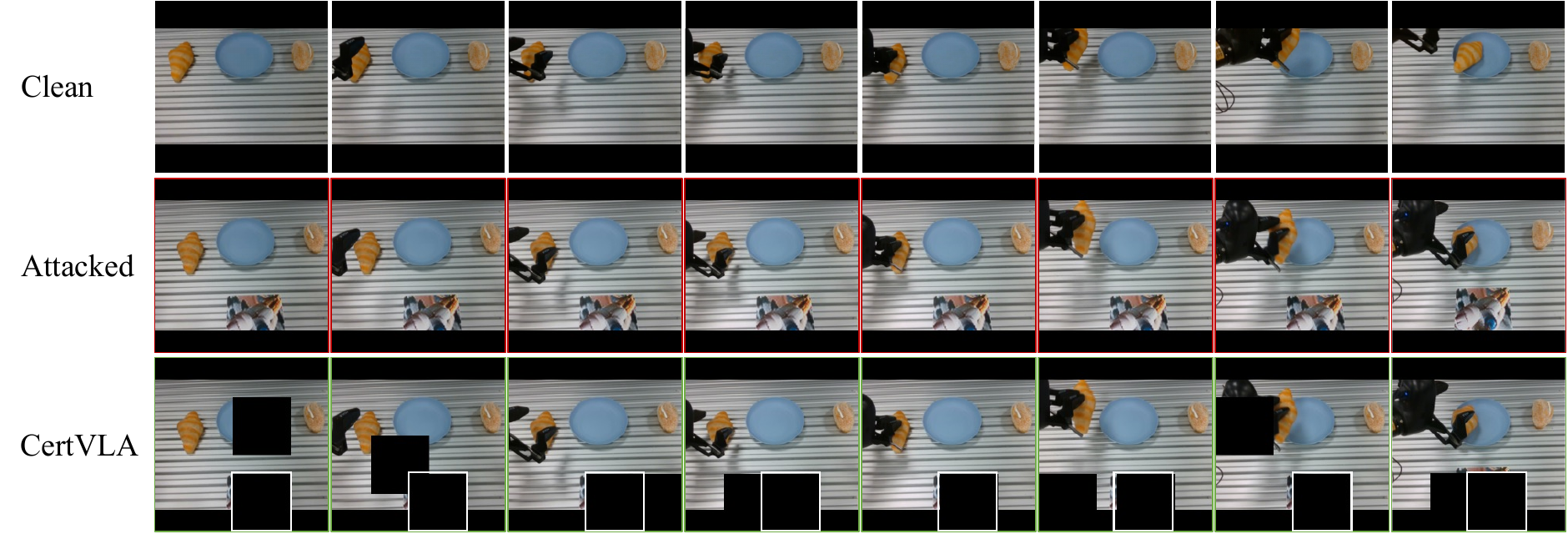}
    \caption{Real-robot rollouts across successive policy queries. The bottom row is
    diagnostic: it visualizes the dual-mask candidate with the smallest
    normalized score $z$ at each query. The white outline identifies the
    anchor mask of the accepted row.
    }
    \label{fig:real}
\end{figure*}
    
\begin{figure*}[t]
    \centering
    \includegraphics[width=0.87\linewidth]{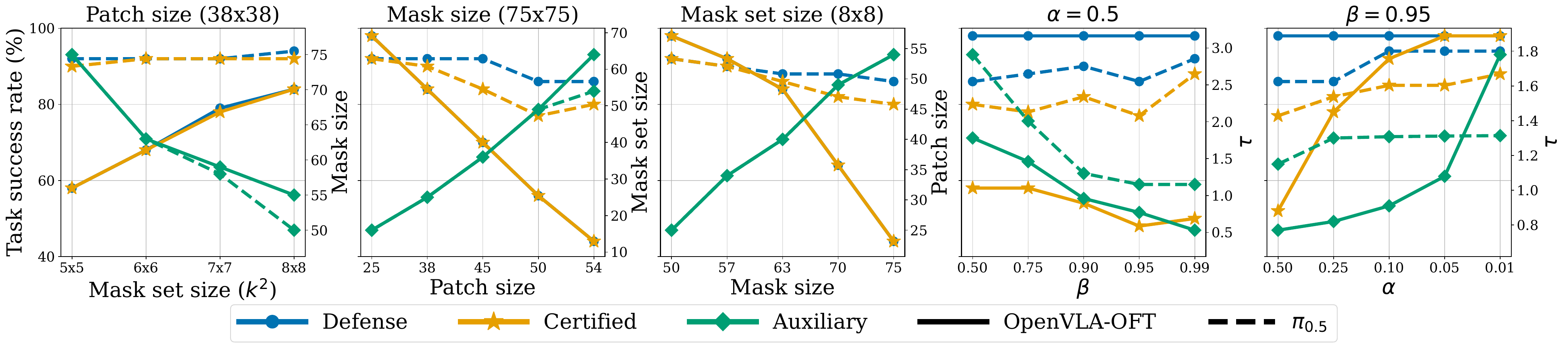}
    \caption{Sensitivity to mask-set size, certified patch size, mask size,
    pair-scale quantile $\beta$, and conformal level $\alpha$ (left to right). Blue and orange report
    Defense and Certified task success; green reports the corresponding
    auxiliary quantity on the right axis.}
    \label{fig:ana}
\end{figure*}
\subsection{Main Results} 
\paragraph{Defense against patch \& texture attack in simulation.}
In Tab.~\ref{tab:libero_attack_results}, OpenVLA-OFT obtains $94\%$ average
Defense and Certified success, while OpenVLA$^\dagger$ reaches $94\%$ Defense
and $82.5\%$ Certified success. The larger gaps for the generative policies
show that some successful rollouts contain a query outside the calibrated
region, confirming the value of reporting empirical recovery and
certification separately.
Defense against texture attack shows transfer beyond planar patches.
$\pi_{0.5}$ performs best, averaging $94.5\%$ Defense and $88.5\%$ Certified
success; $\pi_0$ reaches $84\%$ and $62.5\%$, respectively. OpenVLA variants
are less stable on the Long suite. The result matches our support-based
threat model: certification depends on covering the visible corruption, not
on how its texture was generated.

\paragraph{Defense against physical real patch attack.}
\textit{Quantitative results.}
Tab.~\ref{tab:real-patch-results} shows that the patch reduces $\pi_{0.5}$
success from $90\%$ to $40\%$. CertVLA reaches $60\%$, a 20-point gain that
recovers 40\% of the attack-induced loss. Its $30\%$ Certified success means
that half of the successfully defended trials pass at every query. The
remaining Defense-Certified gap identifies successful rollouts with at least
one query outside the calibrated region. It reflects a stricter certificate under physical environment.
\textit{Qualitative analysis.}
Fig.~\ref{fig:real} separates execution from diagnostic checks. Its third row
shows only a reference dual-mask candidate. The one with minimum $z$-score at each
query is illustrated, which does not select the action. Alg.~\ref{alg:dual_mask_recovery}
instead finds an anchor $i$ whose entire row satisfies $z_{i,j}\leq\tau$
($\exists i\,\forall j$). The white box marks this anchor, and $A_i$ is
predicted from an input masking only its pixels. The other black region
visualizes one second-mask check and is absent during execution. Thus, dual
masks establish the certificate while the executed single-mask input retains
more task-relevant content.

\subsection{Discussion}
\noindent \textbf{Covering, utility, and calibration trade-offs.}
Fig.~\ref{fig:ana} analyzes the principal design parameters from left to
right. Let $P$, $M$, and $s$ denote the patch side length, mask side length,
and mask stride. The mask family is $\mathcal R$-covering only if
 $P\leq M-s+1$
so only configurations satisfying this
constraint are considered. Across these valid configurations, $\pi_0$
maintains a consistently high Defense rate, whereas OpenVLA-OFT is more
sensitive to the masking configuration. 
With $P=38$ fixed, increasing the mask set provides denser coverage and permits smaller masks, reducing visual occlusion at the cost of more mask evaluations. This
substantially improves OpenVLA-OFT Defense, while$\pi$ remains consistently high. Certified success also increases for both models, indicating that finer coverage benefits both task utility and certifiable consistency. 
With $M=75$ fixed, larger certified patches require denser mask placement and
a larger mask set to maintain $\mathcal R$-coverage. But both Defense and Certified
rates decrease as $P$ grows, showing that certifying a larger attack region places greater demands on the masked policy. 
With the mask set fixed at $8{\times}8$, increasing $M$ enlarges the occluded
region while keeping the number of candidate masks unchanged. Both Defense and
Certified rates decrease as $M$ grows, indicating that larger masks remove more
task-relevant visual information and make the recovered actions less reliable.
The degradation is stronger for OpenVLA-OFT, whereas $\pi$ remains more stable,
showing greater tolerance to increased occlusion under the same covering
density.

Calibration is also policy dependent. Increasing $\beta$ rescales $Q_{i,j}$
and changes normalized $\tau$ non-monotonically, and we set it to $0.95$ to mitigate the influence of extreme values while preventing an overly large $\tau$ and an excessively loose certificate.
Reducing $\alpha$ raises acceptance for both policies but loosens the consistency region (Lemma~\ref{lem:conformal}). Neither
parameter replaces $\mathcal R$-covering for deterministic patch erasure.

\noindent\textbf{Interpreting the certification gap.}
Defense success measures if the recovered policy completes a task,
whereas Certified success additionally requires every executed query to
satisfy the calibrated consistency test. A rollout may therefore succeed yet
remain uncertified because of one ambiguous observation. Small gaps indicate
stable masked actions throughout an episode, while larger gaps expose queries that
may gain from improved masked-model utility or calibration.




\section{Conclusion}
CertVLA provides a certified defense for continuous closed-loop VLA control
by combining position-normalized action
consistency and episode-level calibration. Our analysis establishes
query- and rollout-level guarantees that relate certified actions to
attack-erased predictions under bounded-support attacks, with task success
further guaranteed when dual-mask rollout correctness holds. Experiments demonstrate both empirical defense effectiveness
and certifiable robustness across VLA models and physical attacks. 

\nocite{yu2025towards, Li_Ji_Wu_Li_Qin_Wei_Zimmermann_2024, 10.1145/3581783.3611847,Li_2025_CVPR,li2025secureondevicevideoood, li-etal-2025-treble, yu2024unlearnable, yang2025vidlbeval, li2025personalizedconversationalbenchmarksimulating, liang2023efficient,  lu2025pretrain, guo2022unified, guo2025scan, DBLP:journals/corr/abs-2505-05279, yu2025backdoor, lu2026universal}
\bibliography{aaai2027}

\end{document}

%% file: sections/certvla_core.tex
\begin{figure*}[t]
    \centering
    \includegraphics[width=0.85\linewidth]{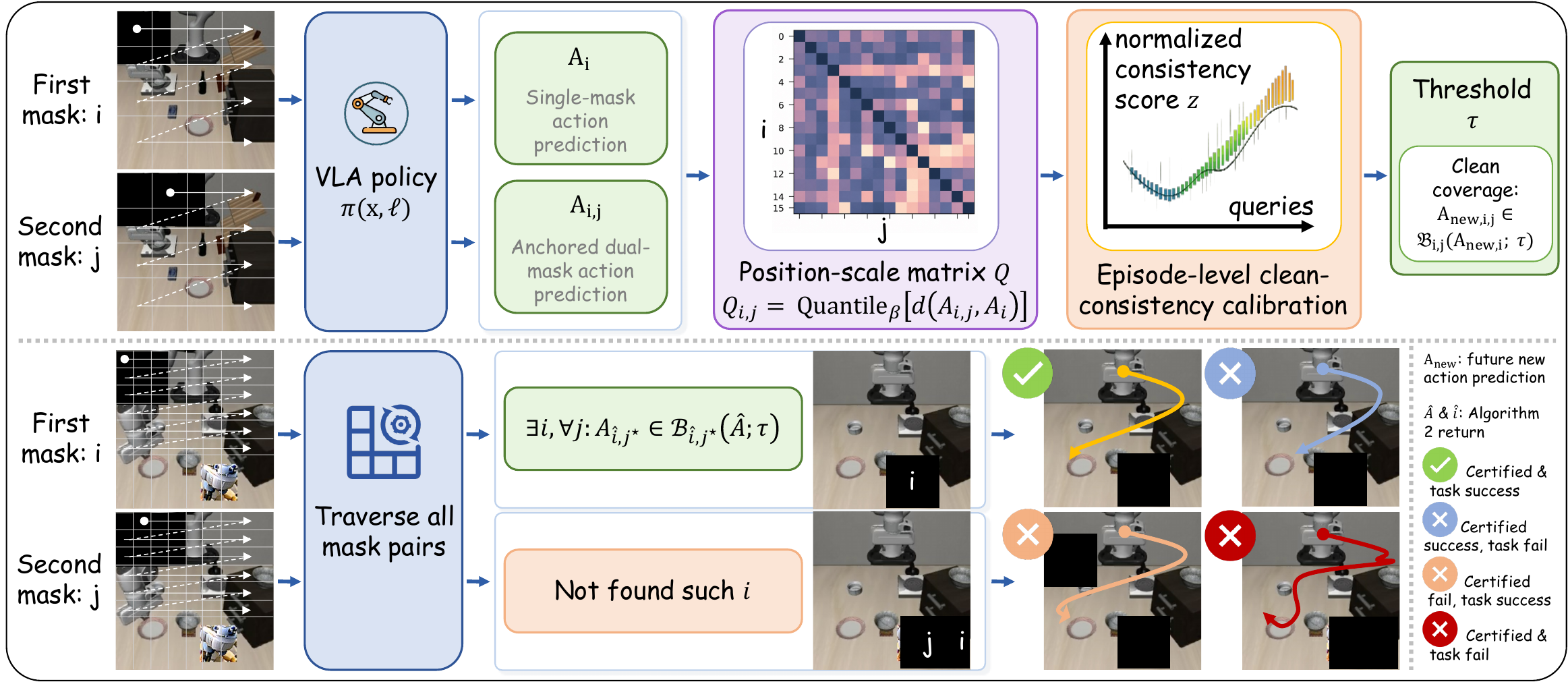}
    \caption{CertVLA framework. \emph{Top:} clean single- and dual-mask
    predictions estimate the position-scale matrix $\mathbf Q$ and calibrate
    threshold $\tau$. \emph{Bottom:} at each query, CertVLA returns a
    single-mask anchor only if it is consistent with every second-mask
    prediction, then composes query flags into an episode certificate distinct
    from empirical task success.}
    \label{fig:framework}
\end{figure*}

\section{Related Work}

\paragraph{VLA models.}
VLA policies map vision and language directly to robot actions. Early scaling were established by RT-1, PaLM-E, and RT-2
~\cite{brohan2022rt1,driess2023palme,zitkovich2023rt2}. RoboFlamingo adapts
open VLMs through imitation learning, whereas Open X-Embodiment and Octo
exploit heterogeneous cross-robot data
~\cite{li2023roboflamingo,openx2023,octo2024}. OpenVLA provides an open 7B
policy and OpenVLA-OFT improves action generation through parallel
decoding and action chunking~\cite{kim2025openvla,kim2025fine}. $\pi_0$ and $\pi_{0.5}$ policies use flow matching and heterogeneous
co-training for dexterous, open-world control
~\cite{black2024pi0,black2025pi05}. Recent VLAs continue to improve generalization, cross-embodiment control, and
closed-loop manipulation through larger-scale training, experience-driven
learning, and active perception
~\cite{intelligence2025pi,liu2026activevla}, while their
certifiable robustness to physical adversarial perturbations remains
largely unexplored.

\noindent\textbf{Adversarial attacks \& defense in robotics.}
Printable patches remain effective across physical transformations
~\cite{brown2017patch}. RoboticAttack demonstrates untargeted,
trajectory-targeted, and physical attacks on VLA control, while UPA-RFAS
transfers patches across architectures, tasks, viewpoints, and sim-to-real
settings~\cite{wang2025vlaattack,lu2026robots}. Tex3D further turns 3D object
textures into persistent attack surfaces~\cite{chen2026tex3d}. Recent defenses
improve empirical stability to visual or multimodal perturbations
~\cite{guo2025robustvla,xie2026strongvla,fu2026stablevla}, but their observed
robustness does not provide a per-rollout guarantee against adaptive attacks.
CertVLA instead certifies any visible corruption whose projected support
satisfies the bounded-region threat model.

\paragraph{Certified defenses against patch attacks.}
Patch certificates use interval bounds or deterministic ablation
~\cite{chiang2020certified,levine2020derandomized}, restricted receptive
fields and masking~\cite{xiang2021patchguard,xiang2021patchguardpp}, or
occlusion consistency~\cite{mccoyd2020minority,xiang2022patchcleanser}.
ScaleCert improves high-resolution certification, PatchCURE exposes tunable
robustness--utility--efficiency trade-offs, and CertMask reduces redundant
mask evaluations~\cite{han2021scalecert,xiang2024patchcure,lyu2026certmask}.
PatchDEMUX extends single-label certificates to multi-label prediction
~\cite{jacob2025patchdemux}. Nevertheless, these guarantees rely on discrete
label or detection decisions. Exact agreement is unsuitable for continuous,
temporally correlated action chunks, and an isolated prediction certificate
does not cover states reached later in a rollout. CertVLA addresses both gaps
through calibrated anchored consistency and episode-level composition.

\section{Preliminaries}
\subsection{VLA Policy and Threat Model}
\label{subsec:threat-model}
Let $x_t=(x_t^{\mathrm{ext}},x_t^{\mathrm{wrist}},s_t)$ contain the
external-camera image, wrist-camera image, and proprioceptive state at policy
query index $t$, respectively, and let $\ell$ be a language instruction. A
VLA policy $\pi$ produces
an action chunk
\begin{equation} \small
 A_t=\pi(x_t,\ell)\in\mathcal A\subseteq\mathbb R^{H\times D},
\end{equation}
of horizon $H$ and action dimension $D$. The controller executes only the
first $h\leq H$ actions before next query.

The adversary may arbitrarily replace pixels in a single connected region
$\Omega_t$ whose image-plane footprint belongs to a known family
$\mathcal R(P)$, where $P$ is the certified maximum side length and
$\mathcal R(P)$ contains every axis-aligned region no larger than $P\times P$ pixels.
For either camera view,
\begin{equation} \small
 x_t^{\Omega}=(1-\mathbf 1_{\Omega_t})\odot x_t
 +\mathbf 1_{\Omega_t}\odot\delta_t .
\end{equation}
Here $\mathbf 1_{\Omega_t}$ is the binary indicator of the attacked support,
$\delta_t$ is arbitrary adversarial pixel content, and $\odot$ denotes
elementwise multiplication. Superscript $\Omega$ denotes the resulting attacked
observation. We make no assumptions about the patch content, generation method,
location, or physical transformation. Our certification only requires that its
projected support in each defended view lies within a region in
$\mathcal R(P)$.

\subsection{Covering Masks and Action Predictions}
The patch location is unknown at inference time, so no single predetermined
mask can guarantee removal. Following PatchCleanser~\cite{xiang2022patchcleanser},
we enumerate $K$ deterministic masks, where index
$i\in\{1,\ldots,K\}$ and operator $M_i$ overwrites a fixed region with a benign
value. The covering property guarantees that, for any admissible bounded patch, the candidate set contains at least one mask that completely removes its support, independent of the patch appearance.
\begin{definition}[$\mathcal R$-covering family]
A deterministic family $\mathcal M=\{M_1,\ldots,M_K\}$ is
$\mathcal R$-covering if, for every $\Omega\in\mathcal R(P)$, some mask
$M_{j^\star}$ overwrites all pixels in $\Omega$.
\end{definition}
\vspace{1mm}
For a query $(x,\ell)$, we define the unmasked, single-mask, and anchored dual-mask predictions
\begin{equation} \small
 \begin{aligned}
 \!\!A_0\!=\!\pi(x,\ell), ~A_i\!=\!\pi(M_i(x),\ell),
 ~A_{i,j}\!=\!\pi(M_j(M_i(x)),\ell).
 \end{aligned}
\end{equation}
For an attacked input $x^\Omega$, superscript $\Omega$ denotes the analogous
predictions $A_i^\Omega=\pi(M_i(x^\Omega),\ell)$ and
$A_{i,j}^\Omega=\pi(M_j(M_i(x^\Omega)),\ell)$. This notation separates the
attacked quantities observed at deployment from their clean counterparts.
The single-mask prediction $A_i$, rather than every pair of dual-mask
predictions, will serve as the anchor of a candidate recovery row.

\section{Methodology}
\textbf{Overview.} Fig.~\ref{fig:framework} summarizes CertVLA's calibration and deployment
stages. On clean episodes, CertVLA normalizes mask-dependent action variation
and calibrates an episode-level threshold. At deployment, it returns a
single-mask anchor only when the anchor is consistent with every second-mask
prediction, and certifies the rollout only if every policy query passes.
The covering property then ties each certified action to a clean prediction
from which the attack is erased.

\subsection{Continuous-Action Consistency Score}
Unlike discrete labels~\cite{xiang2022patchcleanser,lyu2026certmask},
behaviorally equivalent action chunks need not match exactly. We therefore
measure normalized action deviation and account for the distinct benign scale
of each ordered mask pair.

\paragraph{Directional action distance.}
Let $[a_u^-,a_u^+]$ be the valid range of action coordinate
$u\in\{1,\ldots,D\}$ and define
\begin{equation} \small
 \begin{aligned}
 d(A,B)&=\frac{1}{hD}\sum_{r=1}^{h}\sum_{u=1}^{D}
 \frac{|A_{r,u}-B_{r,u}|}{\eta_u(B_{r,u})},\\
 \eta_u(b)&=\max\{|a_u^+-b|,|b-a_u^-|,\epsilon\}.
 \end{aligned}
 \label{eq:nad}
\end{equation}
Here $A,B\in\mathbb R^{H\times D}$ are respectively the dual-mask chunk and
single-mask anchor, $r$ indexes the $h$ executed actions, and $u$ indexes action
coordinates. The denominator normalizes each coordinate by its valid range,
with $\epsilon>0$ ensuring numerical stability. Because it is evaluated at
the anchor $B$, $d(A,B)$ is directional.

\paragraph{Position-scale matrix.}
Mask pairs occlude different scene content and therefore have different clean
deviation scales. Let $e$ index an episode, $q\in\{1,\ldots,N_e^{\rm qry}\}$
a policy query, $N_e^{\rm qry}$ the number of queries, and
$i,j\in\{1,\ldots,K\}$ an ordered mask pair. Let $A_{e,q,i}$ and
$A_{e,q,i,j}$ be the corresponding single- and dual-mask predictions. From
clean scale-calibration episodes $\mathcal C_{\rm scale}$, we collect
\begin{equation} \small
 \begin{aligned}
 \mathcal D_{i,j}^{\rm scale}
 =\{d(A_{e,q,i,j},A_{e,q,i}):\
 &e\in\mathcal C_{\rm scale},
 &q=1,\ldots,N_e^{\rm qry}\},
 \end{aligned}
 \label{eq:pair-scale-set}
\end{equation}
and set its reference scale to the fixed $\beta$-quantile,
\begin{equation} \small
 Q_{i,j}=\widehat{\operatorname{Quantile}}_{\beta}
 \bigl(\mathcal D_{i,j}^{\rm scale}\bigr).
 \label{eq:pair-scale}
\end{equation}
where $\beta\in(0,1)$ is chosen before row calibration. The matrix
$\mathbf Q=[Q_{i,j}]\in\mathbb R_{\geq0}^{K\times K}$ removes pair-specific
scale. 

\begin{algorithm}[t]
\caption{Position-Aware Episode Calibration}
\label{alg:calibrated_threshold}
\begin{algorithmic}[1]
\REQUIRE Scale episodes $\mathcal C_{\rm scale}$; row episodes
$\mathcal C_{\rm row}$; masks $\mathcal M$; quantile $\beta$; clean-episode
miscoverage level $\alpha$; $\epsilon$
\ENSURE Position-scale matrix $\mathbf Q=[Q_{i,j}]$ and threshold $\tau$
\STATE \textbf{\# Stage I: position-scale estimation}
\FOR{$i=1$ to $K$}
  \FOR{$j=1$ to $K$}
    \STATE $\mathcal D_{i,j}^{\rm scale}\leftarrow
    \{d(A_{e,q,i,j},A_{e,q,i}):e\in\mathcal C_{\rm scale},
    q=1,\ldots,N_e^{\rm qry}\}$; ~~$Q_{i,j}\leftarrow
    \widehat{\operatorname{Quantile}}_\beta
    (\mathcal D_{i,j}^{\rm scale})$
  \ENDFOR
\ENDFOR
\STATE \textbf{\# Stage II: episode-level joint calibration}
\FORALL{$e\in\mathcal C_{\rm row}$}
  \STATE $z_{e,q,i,j}\leftarrow d(A_{e,q,i,j},A_{e,q,i})/
  (Q_{i,j}+\epsilon)$ for all $q,i,j$
  \STATE $S_e\leftarrow\max_q\min_i\max_j z_{e,q,i,j}$
\ENDFOR
\STATE $\tau\leftarrow
\widehat Q^{\rm conf}_{1-\alpha}(\{S_e\})$
\RETURN $(\mathbf Q,\tau)$
\end{algorithmic}
\end{algorithm}

\subsection{Position-Aware Clean-Consistency Calibration}
The local scales $\mathbf Q$ make mask pairs comparable but do not set a joint
acceptance threshold. We therefore calibrate an episode-level normalized radius $\tau$, using disjoint scale and row splits to preserve the exchangeability required for conformal calibration. We define normalized consistency score
\begin{equation} \small
 z_{e,q,i,j}
 =\frac{d(A_{e,q,i,j},A_{e,q,i})}{Q_{i,j}+\epsilon},
 \label{eq:pair-score}
\end{equation}
where $\epsilon$ handles zero empirical variation. Thus $z_{e,q,i,j}$ measures
deviation relative to the clean behavior of the same ordered pair, enabling a
shared threshold across positions.

\begin{definition}[Pair-specific clean-consistency region]
\label{def:clean-consistency-region}
For an ordered mask pair $(i,j)$, anchor action $B\in\mathcal A$, and
tolerance $\tau\geq0$, define
\begin{equation} \small
 \mathcal B_{i,j}(B;\tau)
 :=
 \{A\in\mathcal A:
 d(A,B)\leq\tau(Q_{i,j}+\epsilon)\}.
 \label{eq:clean-consistency-region}
\end{equation}
We call $A$ $(i,j,\tau)$-consistent with $B$ when
$A\in\mathcal B_{i,j}(B;\tau)$. Here, $\tau(Q_{i,j}+\epsilon)$ is the raw tolerance for pair $(i,j)$.
\end{definition}
\vspace{1mm}

\begin{lemma}[Score-region equivalence]
\label{lem:score-region}
For every episode $e$, policy query $q$, ordered mask pair $(i,j)$, and
tolerance $\tau\geq0$,
\begin{equation} \small
 z_{e,q,i,j}\leq\tau
 \quad\Longleftrightarrow\quad
 A_{e,q,i,j}\in
 \mathcal B_{i,j}(A_{e,q,i};\tau).
 \label{eq:pair-consistency-region}
\end{equation}
\end{lemma}
\vspace{1mm}

The equivalence follows from $Q_{i,j}+\epsilon>0$.
Stage II calibrates the event dictated by $\mathcal R$-covering: at each query,
some row must be consistent for every second mask, including the unknown mask
that erases the patch:
\begin{equation} \small
 \mathsf{Pass}_e(\tau)
 :=\bigl[\forall q\;\exists i\;\forall j:
 A_{e,q,i,j}\in\mathcal B_{i,j}(A_{e,q,i};\tau)\bigr].
 \label{eq:logical-acceptance}
\end{equation}
Its scalar score is
\begin{equation} \small
 S_e=\max_q\min_i\max_j z_{e,q,i,j}.
 \label{eq:episode-score}
\end{equation}
Here $\max_j$, $\min_i$, and $\max_q$ implement $\forall j$, $\exists i$, and
$\forall q$, respectively. 
Consequently, $S_e$ is the minimum threshold required for every policy query in episode $e$ to admit at least one row whose normalized disagreement remains within the threshold for all second-mask positions.
\begin{lemma}[Score-quantifier equivalence]
\label{lem:score-quantifiers}
For all episode $e$ and threshold $\tau\geq0$, we have (proof in  \textit{Supplementary}.):
\begin{equation} \small
 \begin{aligned}
 S_e\leq\tau
\Longleftrightarrow \mathsf{Pass}_e(\tau) \Longleftrightarrow
 \forall q\;\exists i\;\forall j:\ z_{e,q,i,j}\leq\tau.
 \end{aligned}
 \label{eq:score-quantifiers}
\end{equation}
\end{lemma}
\vspace{1mm}

\noindent\textbf{Episode-level conformal calibration.}
Although $\mathbf Q$ normalizes the location-dependent disagreement of
individual mask pairs, the aggregated score $S_e$ can still vary across clean
episodes due to variations in initial states, object configurations, executed
trajectories, and policy-query sequences. We therefore calibrate the global
acceptance threshold at the episode level rather than fixing it heuristically.

Let $\alpha\in(0,1)$ denote the target clean-episode miscoverage rate. Given
$n=|\mathcal C_{\rm row}|$ calibration episodes with ordered scores
$S_{(1)}\leq\cdots\leq S_{(n)}$, we set
\begin{equation} \small
    \tau
    =
    S_{(k)},
    \qquad
    k=
    \left\lceil
    (n+1)(1-\alpha)
    \right\rceil,
    \label{eq:conformal-quantile}
\end{equation}
provided that $k\leq n$.
Under the standard exchangeability assumption between the calibration
episodes and a future clean episode, this conformal quantile provides
finite-sample marginal coverage
\begin{equation} \small
    \Pr(S_{\rm new}\leq\tau)\geq 1-\alpha.
\end{equation}
Equivalently, a new clean episode is rejected because its consistency score
exceeds $\tau$ with probability at most $\alpha$. Thus, $\alpha$ controls the
allowed clean-episode rejection rate, while $\tau$ is the corresponding
data-dependent tolerance for episode-level action consistency. A smaller
$\alpha$ yields a larger $\tau$, reducing clean rejection at the cost of a
looser consistency certificate.

\paragraph{Algorithm design.}
Alg.~\ref{alg:calibrated_threshold} partitions clean, unattacked episodes
into $\mathcal C_{\rm scale}$ and $\mathcal C_{\rm row}$. Stage I estimates
each ordered-pair scale $Q_{i,j}$ by Eq.~\eqref{eq:pair-scale} and freezes
$\mathbf Q$ before Stage II.
At deployment, fixed $Q_{i,j}$ supplies the pair-specific unit and fixed
$\tau$ thresholds the joint score. Together, Lemmas~\ref{lem:score-region}
and~\ref{lem:score-quantifiers} give the exact action-region and episode-level
interpretations.

\begin{lemma}[Finite-sample clean coverage]
\label{lem:conformal}
If the row-calibration episodes and a future new episode are exchangeable,
Alg.~\ref{alg:calibrated_threshold} satisfies
$\Pr[S_{\rm new}\leq\tau]\geq1-\alpha$. Equivalently, with probability at
least $1-\alpha$, the future episode satisfies
$\forall q\,\exists i\,\forall j:
A_{{\rm new},q,i,j}\in
\mathcal B_{i,j}(A_{{\rm new},q,i};\tau)$.
\end{lemma}
\vspace{1mm}

This clean-coverage result is separate from the deterministic patch-erasure certificate. 
See \textit{Supplementary} for the proof.






\begin{algorithm}[t]
\caption{\textsc{DualMaskRecover}}
\label{alg:dual_mask_recovery}
\begin{algorithmic}[1]
\REQUIRE $(x,\ell)$; VLA $\pi$; masks $\mathcal M$; $\mathbf Q$; $\tau$;
$\epsilon$
\ENSURE Action $\widehat A$, selected row $\hat i$, flag $c$
\STATE $R_{\rm best}\leftarrow+\infty$;
$A_{\rm best}\leftarrow\pi(x,\ell)$; $\hat i\leftarrow0$
\FOR{$i=1$ to $K$}
  \STATE $A_i\leftarrow\pi(M_i(x),\ell)$; $R_i\leftarrow0$
  \FOR{$j=1$ to $K$}
    \STATE $A_{i,j}\leftarrow\pi(M_j(M_i(x)),\ell)$
    \STATE $R_i\leftarrow\max\!\left\{R_i,
    \frac{d(A_{i,j},A_i)}{Q_{i,j}+\epsilon}\right\}$
    \IF{$R_i>\tau$} \STATE \textbf{break} \ENDIF
  \ENDFOR
  \IF{$R_i<R_{\rm best}$}
    \STATE $(R_{\rm best},A_{\rm best},\hat i)\leftarrow(R_i,A_i,i)$
  \ENDIF
  \IF{$R_i\leq\tau$} \RETURN $(A_i,i,\TRUE)$ \ENDIF
\ENDFOR
\RETURN $(A_{\rm best},\hat i,\FALSE)$
\end{algorithmic}
\end{algorithm}
\subsection{Anchored Dual-Mask Recovery}
The patch location is unknown, so each first-mask prediction serves as an anchor and is checked against every second mask. By $\mathcal R$-covering, at least one check erases the patch.

\begin{lemma}[Attack erasure]
\label{lem:erasure}
Let $x^\Omega$ differ from $x$ only on $\Omega\in\mathcal R(P)$, and let each
mask overwrite its support with fixed values. If $\mathcal M$ is
$\mathcal R$-covering, some $M_{j^\star}$ covers $\Omega$ and, for every
$M_i\in\mathcal M$,
\begin{equation} \small
 M_{j^\star}(M_i(x^\Omega))=M_{j^\star}(M_i(x)).
 \label{eq:patch-erasure}
\end{equation}
\end{lemma}

Alg.~\ref{alg:dual_mask_recovery} updates a running row score as second
masks are evaluated. Because this score is nondecreasing, a row can be safely
discarded as soon as it exceeds $\tau$. A row that survives all $K$ checks has
the complete score
\begin{equation} \small
 R_i=\max_{1\leq j\leq K}
 \tfrac{d(A_{i,j},A_i)}{Q_{i,j}+\epsilon}.
 \label{eq:recovery-row-score}
\end{equation}
The algorithm immediately returns the first complete row satisfying
$R_i\leq\tau$. Hence $c=\mathrm{true}$ only after verifying every second mask,
whereas early stopping merely avoids unnecessary evaluations of a rejected
row. If no row passes, it returns the anchor with the smallest observed
partial score as an explicitly uncertified fallback.

\begin{algorithm}[t]
\caption{Defended Rollout Evaluation}
\label{alg:defended_rollout}
\begin{algorithmic}[1]
\REQUIRE Environment $\mathcal E$; instruction $\ell$; VLA $\pi$; masks
$\mathcal M$; $\mathbf Q$; $\tau$; $\epsilon$; horizon $T$
\ENSURE Task outcome $S_{\rm def}$ and episode flag $C$
\STATE Reset $\mathcal E$; $C\leftarrow\TRUE$; $S_{\rm def}\leftarrow\FALSE$
\FOR{$t=0$ to $T-1$}
  \IF{a new policy query is required}
    \STATE $(\widetilde A_t,\hat i_t,c_t)\leftarrow
    \textsc{DualMaskRecover}(x_t,\ell,\pi,\mathcal M,$
    \STATE \hspace{2em}$\mathbf Q,\tau, \epsilon)$; ~~$C\leftarrow C\land c_t$
  \ENDIF
  \STATE Execute the next action from $\widetilde A_t$ in $\mathcal E$
  \IF{task success} \STATE $S_{\rm def}\leftarrow\TRUE$; \textbf{break} \ENDIF
\ENDFOR
\RETURN $(S_{\rm def},C)$
\end{algorithmic}
\end{algorithm}


\begin{theorem}[Attack-erased consistency]
Let $x^\Omega$ be any attacked input with $\Omega\in\mathcal R(P)$. If
Alg.~\ref{alg:dual_mask_recovery} returns
$(\widehat A,\hat i,c)$ with $c=\mathrm{true}$, then
$\widehat A=A_{\hat i}^{\Omega}$ and there exists a clean dual-masked
prediction $A_{\hat i,j^\star}$ such that
\begin{equation} \small
 A_{\hat i,j^\star}
 \in\mathcal B_{\hat i,j^\star}(\widehat A;\tau),
 \label{eq:region-certificate}
\end{equation}
or, equivalently,
\begin{equation} \small
 d(A_{\hat i,j^\star},\widehat A)
 \leq \tau(Q_{\hat i,j^\star}+\epsilon)
 \leq \tau(Q_{\hat i}^{\max}+\epsilon),
 \label{eq:consistency-certificate}
\end{equation}
where $M_{j^\star}$ covers $\Omega$ and
$Q_{\hat i}^{\max}=\max_jQ_{\hat i,j}$. The bound is independent of the
patch content, shape, attack-generation method, and physical transformation, provided that the attack support
$\Omega\in\mathcal R(P)$ satisfies the certified size bound and can therefore
be fully covered by at least one mask in $\mathcal M$.
\label{thm:certificate}
\end{theorem}
\vspace{1mm}


See \textit{Supplementary} for proof. Notably, Theorem~\ref{thm:certificate} certifies action consistency,
not unconditional task success. For practical utility, the patch-erased prediction
should also preserve task-relevant behavior. \textit{Supplementary} empirically validates this property across VLA models under random single- and dual-masking.

\subsection{Closed-Loop Episode Certificate}
Action consistency certification alone does not imply terminal task success. Following
PatchCleanser's two-mask correctness~\cite{xiang2022patchcleanser}, we state
the required closed-loop condition for continuous actions.

\begin{definition}[Dual-mask rollout correctness]
\label{def:rollout-correctness}
Let $\mathcal T_{\rm qry}$ denote the policy-query times of a defended rollout,
and let $\widetilde A_t$ be the recovered action chunk executed at query time
$t$. A task, policy, and environment satisfy \emph{dual-mask rollout
correctness} if every defended rollout for which
\begin{equation}
\forall t\in\mathcal T_{\rm qry},\quad
\exists (i_t,j_t):
\quad
A^{\rm clean}_{t,i_t,j_t}
\in
\mathcal B_{i_t,j_t}(\widetilde A_t;\tau)
\label{eq:rollout-correctness}
\end{equation}
reaches the task-success set.
Here $A^{\rm clean}_{t,i_t,j_t}$ denotes the patch-erased dual-mask prediction
evaluated at the same reached state as $\widetilde A_t$.
\end{definition}
\vspace{1mm}

\begin{table*}[t]
    \centering
    \small
    \setlength{\tabcolsep}{3pt}
    \begin{tabularx}{\textwidth}{
        @{}
        >{\centering\arraybackslash}p{2.15cm}
        l
        *{10}{>{\centering\arraybackslash}X}
        @{}
    }
        \toprule
        \multirow{2}{*}{Attack Type $\downarrow$}& \multirow{2}{*}{Model $\downarrow$}
        & \multicolumn{2}{c}{LIBERO-Spatial}
        & \multicolumn{2}{c}{LIBERO-Object}
        & \multicolumn{2}{c}{LIBERO-Goal}
        & \multicolumn{2}{c}{LIBERO-Long}
        & \multicolumn{2}{c}{Average} \\
        \cmidrule(lr){3-4}
        \cmidrule(lr){5-6}
        \cmidrule(lr){7-8}
        \cmidrule(lr){9-10}
        \cmidrule(l){11-12}

         &
        & Defense & Certified
        & Defense & Certified
        & Defense & Certified
        & Defense & Certified
        & Defense & Certified \\
        \midrule

        \multirow{4}{2.15cm}{\centering
            Patch attack\\
            \cite{wang2025vlaattack,lu2026robots}}
        & OpenVLA$^\dagger$
        & 96 & 96
        & 96 & 64
        & 96 & 94
        & 88 & 76
        & 94.00 & 82.50 \\

        & OpenVLA-OFT
        & 98 & 98
        & 94 & 94
        & 98 & 98
        & 86 & 86
        & 94.00 & 94.00 \\

        & $\pi_0$
        & 82 & 36
        & 92 & 84
        & 86 & 61
        & 74 & 66
        & 83.50 & 61.75 \\

        & $\pi_{0.5}$
        & 96 & 72
        & 97 & 89
        & 84 & 68
        & 86 & 77
        & 90.75 & 76.50 \\

        \midrule

        \multirow{4}{2.15cm}{\centering
            Texture attack\\
            \cite{chen2026tex3d}}
        & OpenVLA$^\dagger$
        & 84 & 82
        & 77 & 30
        & 83 & 83
        & 37 & 30
        & 70.25 & 56.25 \\

        & OpenVLA-OFT
        & 88 & 88
        & 94 & 92
        & 90 & 90
        & 40 & 37
        & 78.00 & 76.75 \\

        & $\pi_0$
        & 88 & 38
        & 92 & 80
        & 82 & 62
        & 74 & 70
        & 84.00 & 62.50 \\

        & $\pi_{0.5}$
        & 100 & 92
        & 94 & 88
        & 94 & 88
        & 90 & 86
        & 94.50 & 88.50 \\

        \bottomrule
    \end{tabularx}

    \caption{Defense and certified task success rates (\%) against physical
    patch and adversarial object texture attacks on LIBERO.
    Average denotes the mean performance across the four LIBERO suites.
    $\dagger$ denotes our dual-mask fine-tuned model, and fine-tuning details
    are provided in Appendix~\ref{app:finetuning}.}
    \label{tab:libero_attack_results}
\end{table*}


At each $t\in\mathcal T_{\rm qry}$, Alg.~\ref{alg:defended_rollout} invokes
Alg.~\ref{alg:dual_mask_recovery}, executes from the recovered chunk
$\widetilde A_t$, and accumulates
$C=\bigwedge_{t\in\mathcal T_{\rm qry}}c_t$. Because
Eq.~\eqref{eq:nad} jointly scores its action chunk, aggregation is per query. The outputs separate defended success $S_{\rm def}$, episode certification
$C$, and certified success $S_{\rm def}\land C$, and any uncertified fallback
makes $C$ false.

\begin{theorem}[Closed-loop consistency certificate]
Suppose every query's patch support belongs to $\mathcal R(P)$ and $\mathcal M$
is $\mathcal R$-covering. If Alg.~\ref{alg:defended_rollout} returns
$C=\mathrm{true}$, then at every $t\in\mathcal T_{\rm qry}$ there exist indices
$(i_t,j_t^\star)$ and an attack-erased clean prediction satisfying
\begin{equation} \small
 A^{\rm clean}_{t,i_t,j_t^\star}
 \in\mathcal B_{i_t,j_t^\star}(\widetilde A_t;\tau).
 \label{eq:episode-consistency}
\end{equation}
This remains valid if the admissible patch location changes between queries.
\label{thm:episode-consistency}
\end{theorem}
\vspace{1mm}

Theorem~\ref{thm:episode-consistency} certifies the observable action sequence, and
the following lemma states when this consistency also guarantees task success.

\begin{lemma}[Conditional closed-loop task certificate]
Under Theorem~\ref{thm:episode-consistency}, if the task also satisfies
Definition~\ref{def:rollout-correctness}, then $C=\mathrm{true}$ implies
$S_{\rm def}=\mathrm{true}$ for every admissible patch sequence.
\label{lem:rollout-correctness}
\end{lemma}
\vspace{1mm}

Together, the theorem and lemma provide a layered certificate tailored to
closed-loop continuous control: CertVLA first verifies an attacker-independent
action-consistency property and invokes the explicit rollout-correctness
condition only for the stronger task-success claim. This separation keeps the
guarantee logically transparent while allowing $S_{\rm def}$, $C$, and
$S_{\rm def}\land C$ to be evaluated without conflating empirical success with
certification.